\documentclass[journal,twocolumn,letterpaper]{IEEEJERM}

\usepackage{times,amsmath,epsfig}
\usepackage{fancyhdr}
\usepackage{amsmath}
\usepackage{amsfonts}
\usepackage{amssymb}
\usepackage{array}
\usepackage{graphicx}
\usepackage{url}
\usepackage{subfigure}
\usepackage{bm}
\usepackage{breqn}
\usepackage{xcolor}
\usepackage{soul}
\usepackage{amssymb}
\usepackage[breaklinks=true]{hyperref}
\usepackage{breakcites}
\usepackage{algorithm} 
\usepackage{algpseudocode}

\usepackage{siunitx}

\usepackage{mathtools}

\DeclarePairedDelimiter\norm{\lVert}{\rVert}

\usepackage{tabularx}
\usepackage{multirow}
\usepackage{booktabs}
\usepackage{etoolbox}
\newrobustcmd{\B}{\bfseries}
\usepackage{color}

\begin{document}

\vspace{1 cm}

\title{Self-supervised learning-based general laboratory progress pretrained model for cardiovascular event detection}

\author{Li-Chin Chen$^1$, 
        Kuo-Hsuan Hung$^1$, ~\IEEEmembership{Student Member,~IEEE,}
        Yi-Ju Tseng$^2$, ~\IEEEmembership{Member,~IEEE,}
        Hsin-Yao Wang$^3$, 
        Tse-Min Lu$^{4,5,6}$,
	  Wei-Chieh Huang$^{4,6,7}$,
        Yu Tsao$^1$, ~\IEEEmembership{Senior Member,~IEEE}
}

\markboth{}
{L-C. Chen \MakeLowercase{\textit{et al.}}: Generalized laboratory progress pretrain model for cardiovascular diseases progress forecasting}

\twocolumn[
\begin{@twocolumnfalse}
  
\maketitle


\begin{abstract}
Objective: \textnormal{
Leveraging patient data through machine learning techniques in disease care offers a multitude of substantial benefits. Nonetheless, the inherent nature of patient data poses several challenges. Prevalent cases amass substantial longitudinal data owing to their patient volume and consistent follow-ups, however, longitudinal laboratory data are renowned for their irregularity, temporality, absenteeism, and sparsity; In contrast, recruitment for rare or specific cases is often constrained due to their limited patient size and episodic observations. This study employed self-supervised learning (SSL) to pretrain a generalized laboratory progress (GLP) model that captures the overall progression of six common laboratory markers in prevalent cardiovascular cases, with the intention of transferring this knowledge to aid in the detection of specific cardiovascular event.} Methods and procedures: \textnormal{GLP implemented a two-stage training approach, leveraging the information embedded within interpolated data and amplify the performance of SSL. After GLP pretraining, it is transferred for target vessel revascularization (TVR) detection.} Results: \textnormal{The proposed two-stage training improved the performance of pure SSL, and the transferability of GLP exhibited distinctiveness. After GLP processing, the classification exhibited a notable enhancement, with averaged accuracy rising from 0.63 to 0.90. All evaluated metrics demonstrated substantial superiority (\emph{p} $<$ 0.01) compared to prior GLP processing.} Conclusion: \textnormal{Our study effectively engages in translational engineering by transferring patient progression of cardiovascular laboratory parameters from one patient group to another, transcending the limitations of data availability. The transferability of disease progression optimized the strategies of examinations and treatments, and improves patient prognosis while using commonly available laboratory parameters. The potential for expanding this approach to encompass other diseases holds great promise.}

Clinical impact: \textnormal{Our study effectively transposes patient progression from one cohort to another, surpassing the constraints of episodic observation. The transferability of disease progression contributed to cardiovascular event assessment.}  \end{abstract}


\end{@twocolumnfalse}]

{
  \renewcommand{\thefootnote}{}%
  \footnotetext[1]{$^1$ Research Center for Information Technology Innovation, Academia Sinica, Taipei, Taiwan.}
  \footnotetext[2]{$^2$ Department of Computer Science, National Yang Ming Chiao Tung University, Hsinchu, Taiwan.}
  \footnotetext[3]{$^3$ Department of Laboratory Medicine, Chang Gung Memorial Hospital at Linkou, Taoyuan City, Taiwan.}
  \footnotetext[4]{$^4$ Division of Cardiology, Department of Internal Medicine, Taipei Veterans General Hospital, Taipei, Taiwan.}
  \footnotetext[5]{$^5$ Department of Health Care Center, Taipei Veterans General Hospital, Taipei, Taiwan.}
  \footnotetext[6]{$^6$ Department of Internal Medicine, School of Medicine, College of Medicine, National Yang Ming Chiao Tung University, Taipei, Taiwan.}
  \footnotetext[7]{$^7$ Department of Biomedical Engineering, National Taiwan University, Taipei, Taiwan.}
}
 
%
\IEEEpeerreviewmaketitle

\section{Introduction}
%
%
%
%

\IEEEPARstart{R}{egular} surveillance stands as an imperative facet within cardiovascular disorders management \cite{sattar2020improving}. Laboratory analysis constitutes a vital component, involving multifarious chemical tests that scrutinize blood, urine, or body tissue specimens. These tests gauge the body's response to food intake, medication, and treatment, thus providing crucial insights into disease progression and signaling the need for medication or dietary modifications. For chronic diseases, laboratory results are more meaningful when observed longitudinally rather than episodically. A vast repository of longitudinal data has been accumulated for prevalent diseases such as hypertension (HTN) and diabetes mellitus (DM). Nonetheless, the inherent nature of these data, characterized by irregularity, absenteeism, and sparsity, presents challenges in leveraging their full potential for machine learning applications. Conversely, rare or specific cases are frequently associated with a limited patient population and episodic observations, also impeding the integration of machine learning technology in their progression assessment. This study employed self-supervised learning (SSL) to pretrain a \textbf{g}eneralized \textbf{l}aboratory \textbf{p}rogress (GLP) model. GLP captures the overall progression of common laboratory markers in prevalent cardiovascular cases, with the intention of transferring this knowledge to aid in the detection of target vessel revascularization (TVR) occurrences in patients undergoing percutaneous coronary intervention (PCI).

\subsection*{Challenging nature of laboratory data} 

Diverse methodological approaches can be adopted depending on the distinct characteristics of the population under study. Cross-sectional studies, which capture the patient status at a single time point, provide only a temporary snapshot and preliminary glimpses into future disease progression. In contrast, cohort studies, with their longitudinal observations, offer a more comprehensive understanding of disease development \cite{hulley2013designing}. However, the collection of such data over an extended period can be intricate, time-consuming, and costly, frequently suffered from patient dropouts and incomplete data \cite{HCNN2017,CNN_multifusion2016}. 

Consequently, rare or specific diseases frequently resort to cross-sectional studies due to limited patient size, yielding episodic observations. In the context of prevalent cases, longitudinal data is more readily accessible owing to continuous follow-ups. Nonetheless, these observations heavily rely on patient adherence, insurance regulations, clinical guidelines, and the clinical judgment of physicians. Any disruption to these factors can lead to irregularity and sparsity, where observations may be skipped, or sampled irregularly over a prolonged period \cite{FIDDLE, mimic2020}. Similarly to electronic health records (EHRs), laboratory test records encompass a wealth of abundant and longitudinal patient information, still, notable for their irregularity, temporality, and sparsity, often accompanied by noisy outliers and missing values \cite{HCNN2017,CNN_multifusion2016,mimic2020}. 

\subsection*{Related works}

Machine learning techniques have undergone extensive investigation, facilitating diverse applications that contribute to clinical care through the utilization of EHRs \cite{JBI_review2021,miotto2018deep,diabetes_JBHI}. SSL has recently gained attention due to its ability to derive labels for training data directly from the data itself \cite{SSL_survey2020, self_supervised_time_serious_data, SSL_pretrain, SSL_review, VIME_NEURIPS2020}. This offers a unique opportunity to leverage the vast amounts of data available without relying on quality annotations \cite{breakingNewsSSL2022,CLOCS,SSL_review}. Generally, SSL can be categorized into generative and contrastive learning approaches \cite{NatureBiomedicalEngineering2022_SSL_review, chen2020simple}. Generative models possess the capability to generate new samples from the underlying distribution or recover the original data distribution \cite{SSL_review, SSL_ECG_2023}. For instance, Simone et al. \cite{SSL_ECG_2023} showcased the utilization of a generative adversarial network (GAN) for the synthesis of electrocardiography (ECG) data. Their study achieved remarkable results by generating a wide spectrum of ECG patterns that preserved synchronization and abnormalities. Yoon et al. \cite{VIME_NEURIPS2020} leveraged SSL strategies to impute corrupted values and train on unlabeled data in the domain of genomics and clinical data. Furthermore, Lee et al. \cite{GPT4_medicalsummary} employed GPT-4 to summarize physician-patient conversations and generated clinical notes. 


Meanwhile, contrastive learning aims to capture the relationship between input data and prediction targets, thereby generating a global contextual representation that is shared among samples \cite{SSL_review,CLOCS}. For instance, Kiyasseh et al. \cite{CLOCS} devised a temporal and spatial discriminative approach for ECG analysis, extracting patient-specific representations through leveraging contrastive loss. Zhang et al. \cite{NEURIPS2022_194b8dac} employed contrastive learning to capture the distances between temporal and frequency components, and applied the pretrain representations to various time-series databases, including ECG, human activity recognition, and physical status monitoring. Wickstrøm et al. \cite{WICKSTROM202254} proposed a contrastive framework based on mixing up augmentation for uni- and multivariate time-series data, which was then transferred to ECG classification. Furthermore, Ouyang et al. \cite{diabete_retinopathy2023contrastive_two_stage_train} trained an encoder using unlabeled retinal images through contrastive learning, and subsequently fine-tuned the encoder for the classification of reference and non-reference cases.

Many SSL applications focus on learning from a pretext task and transferring the learned representations to a different domain. This entails transfer learning, a concept that improves a classifier in one domain (pretext task) using more readily obtainable data and then applying this acquired knowledge to another domain (downstream task) \cite{SSL_survey}. For instance, Tang et al. \cite{self_HumanActivityRecognition2021} utilized a teacher-student self-training model to capture information from a large-scale unlabeled dataset of wearable and mobile sensing data, which was subsequently transferred to seven different datasets with varying sensor types, populations, and protocols. Similarly, Spathis et al. \cite{SSL_bisignalrepresentation} used activity accelerometer sensor data as input to forecast heart rate and transferred the learned representations to capture physiologically meaningful and personalized information using linear classifiers.

Nontheless, the aforementioned works assume that signals and information were collected on a regular basis and do not address the issues of irregularity, absenteeism, and sparsity commonly encountered in EHRs. Tipirneni et al. \cite{SSL_timeseries} bypass the irregular, absent, and sparse nature of EHRs by training the model in a SSL manner and mask out unobserved forecasts in the loss function during training. Furthermore, SSL applications have been recognized as challenging in terms of finding effective pretrain tasks \cite{NatureBiomedicalEngineering2022_SSL_review}. Most studies have relied on an empirical trial-and-error approach to identify the most suitable pretext tasks. The transferability of pretrain models has shown mixed results when applied to more specific domains such as medicine \cite{medical_SSL_review}. For instance, Liu et al. \cite{transfer_fail} were unable to successfully transfer ImageNet for detecting lymph node metastasis in pathology images, citing significant domain differences between natural scenes and pathology images as the reason for the transfer failure. Another example is the discussion surrounding language representation models for the biomedical domain \cite{lee2020biobert}. Gu et al. \cite{DomainSpecificLanguageModel} argued that models trained on domain-specific vocabulary outperform those trained on general corpora due to differences in word distributions between general and biomedical corpora .

\subsection*{Cardiovascular diseases and associated risk factors}
Cardiovascular disease stands as one of the leading causes of global mortality. Established traditional risk factors incorporate HTN, DM, and smoking \cite{braunwald1997shattuck}. These risk factors can contribute to endothelial injury, plaque formation, and coronary thrombus formation \cite{fuster1994lewis}, consequently driving the progression of cardiovascular disease. PCI has emerged as a widely employed treatment modality for cardiovascular disease \cite{stone2018percutaneous}. 

Although the implementation of drug-eluting stents (DES) has significantly reduced the incidence of TVR in recent years \cite{giustino2022coronary, serruys2006coronary}, the occurrence of TVR after DES implantation, with an incidence ranging from 3\% to 20\%, remains a prevalent clinical concern \cite{dangas2010stent, nakano2013human, stolker2012repeat}. Hence, the prevention of TVR and the reduction of re-admission rates continue to be significant clinical challenges in the field of cardiovascular medicine following PCI. With respect to the timing of cardiac events, patients undergoing PCI face a risk of subsequent adverse events, including TVR \cite{stone2011prospective, cutlip2007clinical}. However, achieving a consensus on accurate preprocedural risk stratification and prognosis assessment to identify high-risk patients prior to PCI remains an ongoing pursuit.

TVR is associated with complex pathophysiological mechanisms involving lipid metabolic disorders \cite{sun2017trends} and inflammatory processes \cite{ridker2017antiinflammatory}. Several previous studies have explored potential predictive factors linked to a high incidence of TVR based on patient and procedure-related variables \cite{kim2015incidence, kastrati2006predictive, cassese2014incidence}. However, dedicated applications for TVR prediction are yet to be developed. Most studies have focused on identifying comprehensive predictors for TVR or developing prediction models without specifically targeting individualized risks \cite{kastrati2006predictive, yeh2011predicting, stolker2017combining}. The collection of data for comprehensive predictors may introduce burden and complexities when integrating such applications into routine clinical practice and thus warrants careful consideration.

\subsection*{Laboratory markers of cardiovascular diseases}
Earlier studies have indicated that preprocedural parameters are associated with cardiovascular disease. Total cholesterol levels (Chol) and low-density lipoprotein cholesterol (LDL-c) are strongly linked to cardiometabolic diseases and widely accepted in diagnostic practices. Conversely, the plasma level of high-density lipoprotein cholesterol (HDL-c) exhibits an inverse relationship with the risk of cardiovascular diseases \cite{gordon1989high}. Clinical studies have highlighted the connection between circulating white blood cells (WBCs) and cardiovascular outcomes, demonstrating that elevated WBC count increases the short- and long-term risk in patients with acute coronary syndromes (ACSs) \cite{zalokar1981leukocyte}. Hage et al. \cite{hage2009glycaemic} reported that baseline fasting blood glucose (glucose AC) predicts restenosis, suggesting that focusing on glucose reduction rather than solely normalizing glucose levels is more beneficial \cite{hage2009glycaemic}. Additionally, serum uric acid (UA) has been identified as a prognostic cardiovascular biomarker, predicting total and cardiovascular mortality in the context of secondary prevention of coronary artery disease, as demonstrated by the Verona Heart Study \cite{mozzini2021serum}. Furthermore, the National Cholesterol Education Program III (NCEP III) recommends the use of Chol or LDL-c in conjunction with HDL-c (Chol/HDL-c, LDL-c/HDL-c) as markers for screening and treating patients with cardiovascular disease, along with the utilization of the 10-year risk Framingham scoring assessment \cite{expert2001executive}.

\vspace*{1 cm}
\section{Methodology}
To address the challenges previously mentioned, we have devised the following propositions for our work: (1) In order to address the challenges stemming from irregularity, absenteeism, and sparsity within longitudinal observations, we have deployed interpolation and SSL techniques to infer absented data. (2) For patients with limited numbers and episodic observations, we developed a pretrain model specifically tailored to capture the temporal latent representation of prevalent cases and transfer disease progress knowledge to these smaller cohorts. (3) Our work is based on commonly available resources, avoiding the need to initiate new trials for extensive patient data collection. We focused on six laboratory parameters: the Chol/HDL-c ratio, LDL-c, the LDL-c/HDL-c ratio, glucose AC, WBC, and UA. (4) Our work leverages the intercorrelation among cardiometabolic diseases as an indication of pretext and downstream tasks.

Our objective is to construct a two-stage pretraining model that captures the laboratory progress of general cases and utilizes this information to predict cardiac events in another patient group. Prior research has indicated that incorporating a two-step training approach, involving pretraining the model on a domain-general dataset followed by training on domain-specific datasets, yields enhancements in transitioning representations to the downstream task \cite{SSL_pretrain, lee2020biobert, diabete_retinopathy2023contrastive_two_stage_train}. Hence, we propose a two-stage training process: Stage 1 involves learning general laboratory progress information based on interpolated data, followed by Stage 2, where SSL is employed to refine the model's progression representation using non-interpolated data. Subsequently, GLP model is fine-tuned to classify the occurrence of TVR.

The following sections outline the detailed methodology of GLP. Initially, we introduced the interpolation and framing method for longitudinal data. Subsequently, we outlined the design of the GLP model, along with the training algorithm and the design of downstream classifier. We also elucidated the validation methods employed for both the pretext and downstream tasks. Lastly, we provided details regarding the patient recruitment process and the datasets utilized for both tasks.

\subsection*{Interpolation methods}
\begin{figure}
\centerline{\includegraphics[width=\columnwidth]{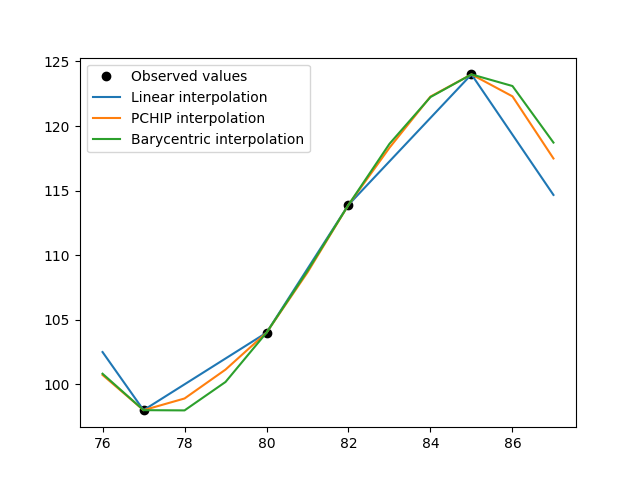}}
\caption{Interpolation results based on different methods. This showcases a segmented period of glucose AC. The x-axis corresponds to the timeline, while the y-axis represents the laboratory values.}
\label{fig_interpolation_difference}
\end{figure}

Interpolation serves the purpose of inferring values that lie between two known observations. It aims to approximate the values of $\hat{f}(x)$ that fulfill the interpolation conditions $\hat{f}(x_{j}) = y_{j}$ for $j = 0, 1, \ldots, n$. This study encompassed the evaluation of three interpolation techniques: linear interpolation, piecewise cubic Hermite interpolating polynomial (PCHIP), and barycentric interpolation. The values in linear interpolation were derived by considering the gradients of the known observations, denoted as:


\begin{equation}
\hat{y}_{j} = y_{i} + (t_{j} - t_{i})\frac{(y_{k}-y_{i})}{(t_{k}- t_{i})},
\label{linear_interpolation}
\end{equation} where $t$ signifies the time of estimation, $\hat{y}_{j}$ denotes the estimated value at $t_{j}$, and $i < j < k$. Meanwhile, PCHIP interpolation \cite{PCHIP} defines $d_{j} = (y_{k}-y_{j})/(t_{k}- t_{j})$ as the slopes at $x_{j}$. If the signs of $d_{j}$ and $d_{i}$ differ or either of them equals zero, $\hat{y}_{j}$ is set to 0. Otherwise, it is determined using the weighted harmonic mean, expressed as:

\begin{equation}
\hat{y}_{j} = \frac{(w_{1}+w_{2})}{\frac{w_{1}}{d_{i}}+\frac{w_{2}}{d_{j}}},
\label{PCHIP_weight}
\end{equation}
where $w_{1} = 2(t_{k} - t_{j}) + (t_{j}-t_{i})$ and $w_{2} = (t_{k} - t_{j}) + 2(t_{j} - t_{i})$. Finally, for barycentric interpolation \cite{2004barycentric,2014barycentric}, a given set of nodes $x_{0}, x_{1}, \ldots, x_{n}$ and masses $w_{0}, w_{1}, \ldots, w_{n}$ are utilized to determine the functions $w_{0}(x), w_{1}(x), \ldots, w_{n}(x)$ satisfying:

\begin{equation}
x = \frac{\sum^{n}_{i=0}w_{i}(x)x_{i}}{\sum^{n}_{i=0}w_{i}(x)}.
\label{barycenter_interpolation}
\end{equation} Here, $x$ represents the barycenter of the nodes, which can be employed for interpolation using:

\begin{equation}
\hat{y} = \sum^{n}_{i=0}b_{i}(x)f_{i},
\label{barycenter_interpolation2}
\end{equation}
where $b_{i}$ corresponds to the linear function. Fig.~\ref{fig_interpolation_difference} illustrates the segmented period of glucose AC, employing different interpolation methods.

\subsection*{Longitudinal data framing}
By defining the laboratory observations for each patient during the study period as $y_{t_{0}}, y_{t_{1}}, \ldots, y_{t_{n}}$, where $t_{i}$ represents the months in the timeline and $t_{i} \in \mathbb{R}$, we designate $y_{t_{0}}$ and $y_{t_{n}}$ as the actual observed values. Let $y_{t_{m}}$ denote the second-to-last observed value of a patient. Interpolation takes place between $y_{t_{0}}$ and $y_{t_{m}}$ when $y_{t_{i}}$ is missing. The observations were organized into a frame with a designated time interval $r$. Therefore, the longitudinal data is framed as $y_{t_{i}}:y_{t_{i+r}}$, with subsequent frames incrementing by one step ($i+1$) while $i+r \leq m-1$. If $\norm{t_{i}, t_{m-1}} \leq r$, the frame is omitted. These segmented frames, denoted as the interpolated data, serve as the input of Stage 1. Their corresponding prediction target is $y_{t_{i+r+1}}$, where $i+r+1 \leq m$.

\begin{figure}
\centerline{\includegraphics[width=\columnwidth]{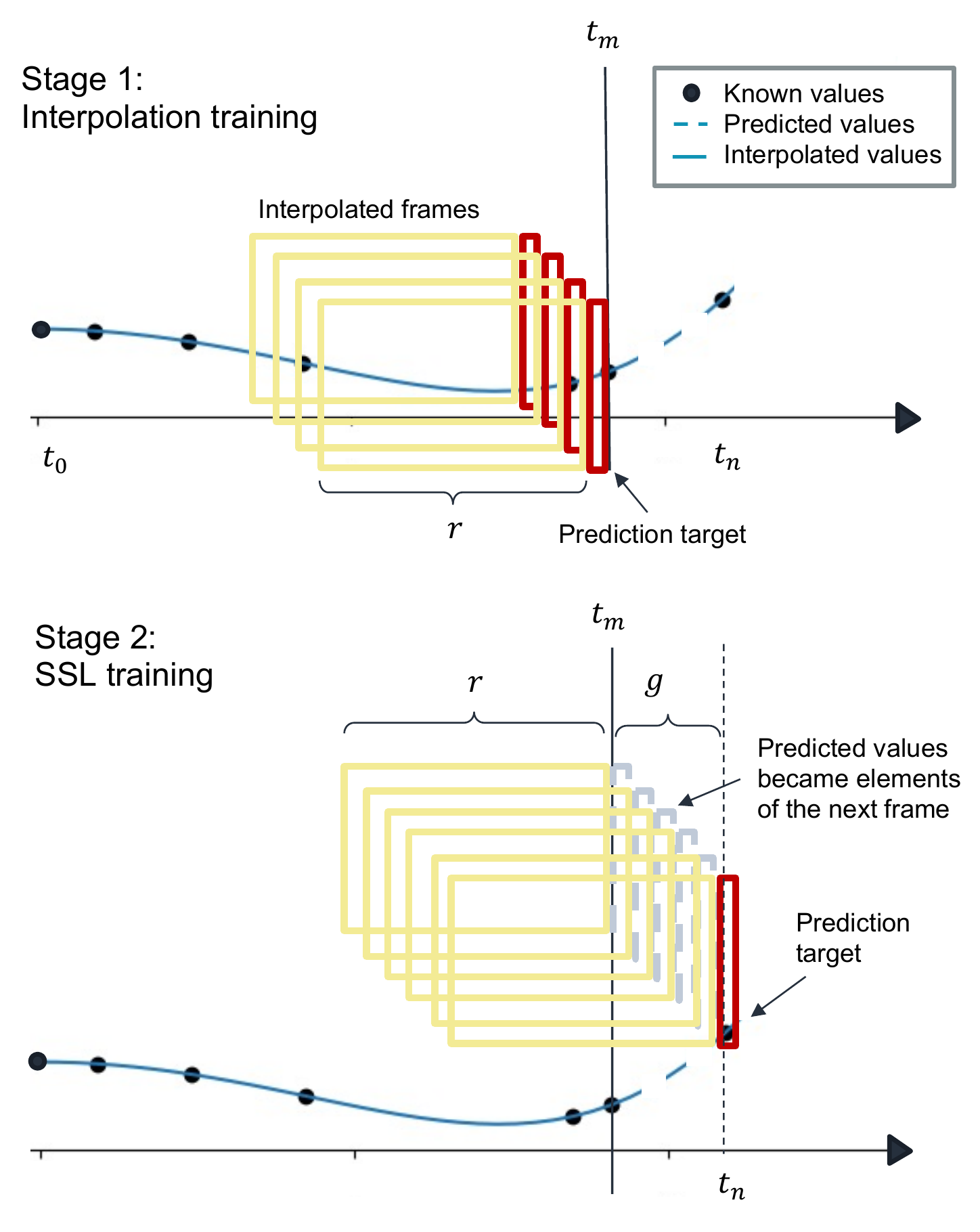}}
\caption{Logitudinal data framing for two-stage training. SSL: self-supervised learning; $r$: time interval (frame size); $g$: time gap between the input frame and prediction target, where $g \leq \frac{r}{2}$.}
\label{fig_framing}
\end{figure}

From the previous stage, the frame $y_{t_{m-r}}:y_{t_{m}}$ and the last observed value $y_{t_{n}}$ are isolated. They are the non-interpolated data in Stage 2, with $y_{t_{m-r}}:y_{t_{m}}$ serving as the input and $y_{n}$ as the prediction target. Considering that coronary stent trials primarily focus on target vessel/lesion-related clinical outcomes within the shorter term, particularly in the first 12 months post-PCI \cite{cutlip2004beyond}, we set $r$ as 12 months. The framing process is visually depicted in Figure~\ref{fig_framing}.

\subsection*{GLP model design}

\begin{figure}
\centerline{\includegraphics[width=0.8\columnwidth]{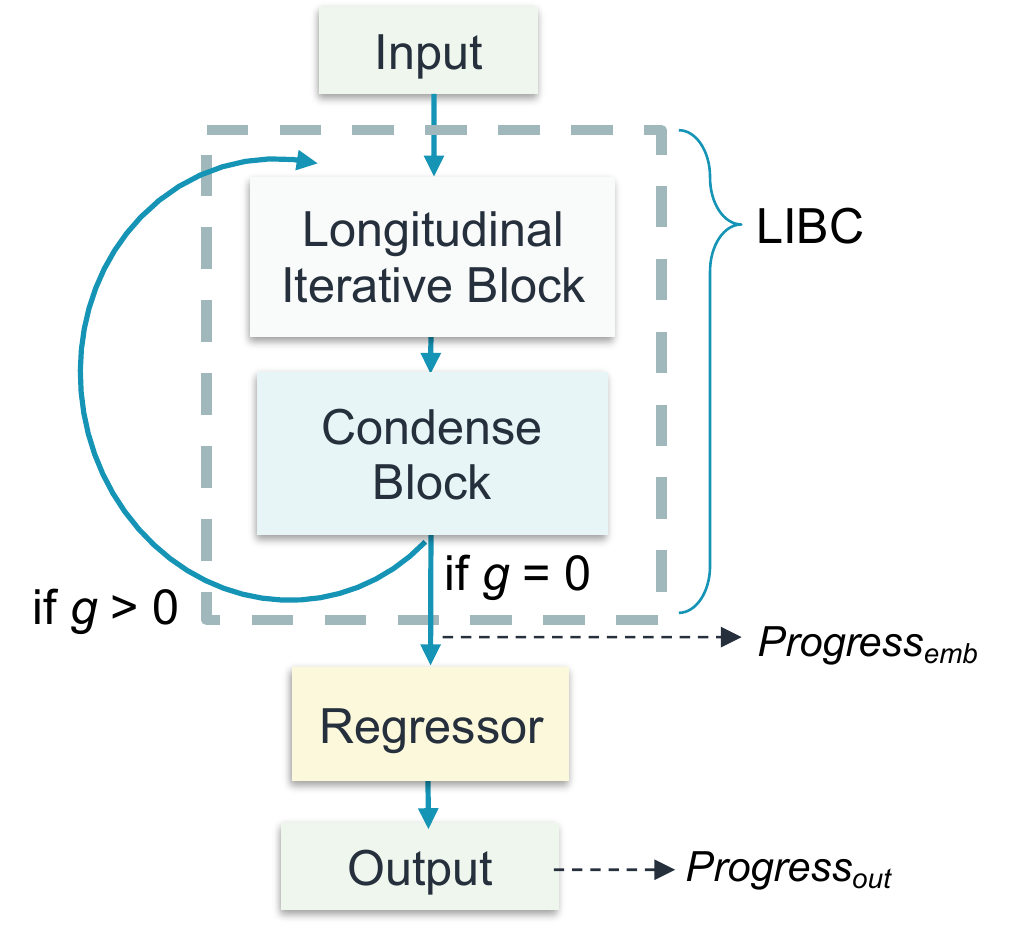}}
\caption{GLP model design. $g$: time gap between the input frame and prediction target; $Progress_{emb}$: latent output of $LIBC$; $Progress_{out}$: latent output of $regressor$.}
\label{fig_model_design}
\end{figure}

GLP was designed to monitor the laboratory progress of patients and forecast their observations for the subsequent month. Fig.~\ref{fig_model_design} presents the model architecture of GLP, which consists of two main components: the longitudinal iterative block ($LIBC$) and a $regressor$. The $LIBC$ comprises a Bidirectional Long Short-Term Memory (BiLSTM) layer and a Fully Connected (FC) condensing layer. A Rectified Linear Unit (ReLU) activation function is applied after each layer to enhance the non-linearity of the model. The BiLSTM processes the input data in both the forward and backward directions \cite{BiLSTM}, enabling the capture of contextual information from past data. The number of hidden nodes for the BiLSTM layer was set to 5. The FC condensing layer is utilized to compress the output of the BiLSTM layer back to the original input, facilitating an autoregressive flow.

On the other hand, the $regressor$ consists of two FC layers followed by a ReLU activation function in between. The number of hidden nodes for the $regressor$ is set as 5, 2, 2, and 1, respectively. The model used the Mean Squared Error (MSE) as loss function, which measures the average squared difference between the predicted values and the actual values.

\subsection*{Two-stage training algorithm}

During Stage 1 training, the interpolated data were utilized in supervised learning. An additional parameter called the number of certainty mask ($certain$) was introduced in Stage 1. It denotes the required number of real observations within a frame, indicating the level of real observations the models require to generate reliable predictions. It also serves as a means to address uncertainty within the training data. Adhering to insurance regulations, patients were scheduled for cardiovascular disease examinations every three months, resulting in a maximum of four actual observations within a 12-month timeframe. Consequently, the range of 0 to 5 was explored for $certain$, and the optimal value was determined as a parameter setting for GLP.

In Stage 2 training, we employed an autoregressive based SSL, which takes inputs from a time series regressed on previous inputs from the same time series. The probability of each input is conditioned on the preceding input, and can be formulated as:

\begin{equation}
\max_{\theta} p_{\theta}(x) = \sum_{t=1}^{T}\log p_{\theta}(x_{t}\mid x_{1:t-1}),
\label{interpolation}
\end{equation}
where $x_{t}$ represents the input at time $t$, $p_{\theta}$ denotes the probability, and $\max_{\theta} p_{\theta}$ signifies the maximized likelihood \cite{SSL_review}. Starting from frame $y_{t_{m-r}}:y_{t_{m}}$, the model utilizes the parameters $\theta$ obtained from Stage 1 and predicts $y_{t_{m+1}}$. The subsequent input frame becomes $y_{t_{m-r+1}}:y_{t_{m+1}}$, with the prediction target located at $y_{t_{m+2}}$. This process continues until the prediction target reached $y_{t_{n}}$. The predicted data then becomes part of the training data for the next frame, allowing the model to learn from its own generation. 

\begin{figure}
\centerline{\includegraphics[width=1.0\columnwidth]{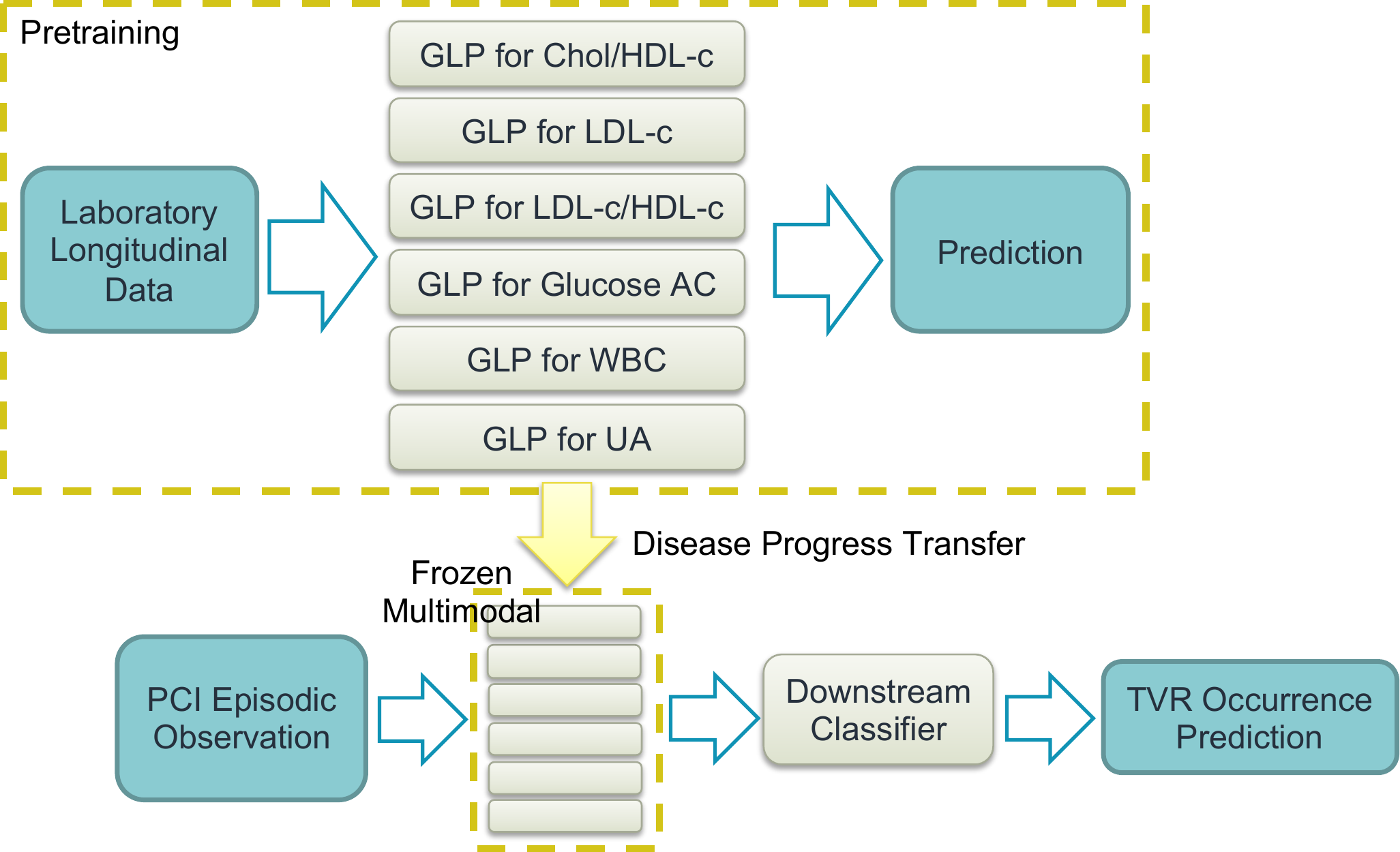}}
\caption{Training pipeline. GLP: general laboratory progress pretrain model; Chol/HDL-c: ratio of total cholesterol and high-density lipoprotein cholesterol; LDL-c: low-density lipoprotein cholesterol; LDL-c/HDL-c: ratio of low-density and high-density lipoprotein cholesterol; Glucose AC: fasting blood sugar; WBC: white blood cells; UA: Uric Acid; PCI: percutaneous coronary intervention; TVR: target vessel revascularization.}
\label{Training_flow}
\end{figure}

To summarize, in Stage 1, there is no time gap ($g = 0$) between the input frame and the prediction target. The input passes through the $LIBC$ once and then enters the $regressor$. On the other hand, in Stage 2, $g = n-m-1$ and $g > 0$, and the process iterates until the prediction target is reached. The training algorithms for Stage 1 and 2 are explicitly outlined in Algorithm \ref{stage1_alg} and \ref{stage2_alg}. Each laboratory parameter was trained individually and optimized to achieve the best performance. The six pretrained GLP models were subsequently concatenated in a multimodal fashion and utilized for domain transfer to perform TVR occurrence classification, as illustrated in Figure~\ref{Training_flow}.

\begin{algorithm}
	\caption{Stage 1 training} 
        \label{stage1_alg}
        \textbf{Input:} \textbf{$X_{interpolated}$}, batch size $N$ 
        \begin{algorithmic}[1]
            \For {$epoch = 0$ to $50$}
                \For {$batch \{x_{k}\}^{N}_{k=1} \in X_{interpolated}$}
                \State pass $x_{k}$ through $LIBC$ and $regressor$ 
                \State Optimize using Adam optimizer
                \EndFor
                \State Save the parameters $\theta$ of the model
            \EndFor
	\end{algorithmic} 
        \textbf{Output:} $\hat{y}$
\end{algorithm}

\begin{algorithm}
	\caption{Stage 2 training} 
        \label{stage2_alg}
        \textbf{Input:} \textbf{$X_{non-interpolated}$}, batch size $N$, time gap between the input frame and prediction target $g$, parameters $\theta$ from stage 1 training
        \begin{algorithmic}[1]
            \For {$epoch = 0$ to $50$}
                \For {$batch \{x_{k}\}^{N}_{k=1} \in X_{non-interpolated}$, $\{g_{k}\}^{N}_{k=1} \in g$}
                    \If {$max(g_{k}) = 0$}
                        \State pass $x_{k}$ through $LIBC_{\theta}$
                        
                     \EndIf
                     \If {$max(g_{k}) > 0$}
                        \For {$t \in \{1, 2, \ldots, max(g_{k})\}$}
                            \State pass $x_{k}$ through $LIBC_{\theta}$
                        \EndFor
                        \For {$g_{i} \in \{1, 2, \ldots, k\}$}
                            \State select estimated $\hat{y_{k}}$ at time stamp $g_{i}+1$
                        \EndFor
                     \EndIf
                     \State pass $\hat{y_{k}}$ through $regressor_{\theta}$
                     \State Optimize using Adam optimizer
                 \EndFor
                \State Update the parameters $\theta$ of the model
            \EndFor
	\end{algorithmic} 
        \textbf{Output:} $\hat{y}$
\end{algorithm}


\subsection*{Input vector and normalization}

The input of GLP consists of five features: age, gender, $certain$, discrete value encoding, and normalized laboratory values. Numeric values, such as age and laboratory values, are normalized using the natural logarithm of one plus the input ($log1p$). This transformation ensures that the values are projected into a vector space above zero, preventing potential errors that could arise from maldistribution between positive and negative values. Additionally, $log1p$ is accurate for small values of $x$, ensuring that $1+x=1$ with floating-point accuracy without significantly altering the original value \cite{log1p}. The use of $log1p$ also prevents information leakage \cite{data_leak}, unlike scaling which necessitates knowledge of the maximum and minimum values.

One-hot encoding is employed for gender, certainty mask, and discrete value encoding. Gender is binary-encoded, with a value of 1 assigned to male and 0 assigned to female. The $certain$ is also binary-encoded. It indicated whether the value is a real observation or an estimation obtained through interpolation (true observation/estimated value). Discrete value encoding categorizes laboratory results into two groups (low and high) or three groups (low, normal, and high) based on different threshold values (summarized in Appendix~\ref{lab_descret}).

\subsection*{GLP validation}
The validation of GLP was conducted by forecasting $y_{n}$ without interpolation support within $\frac{r}{2}$. To further assess the performance of the proposed two-stage training process, we used SSL as baseline (Stage 2 training), and compared with the following approaches: (1) Supervised training based on interpolated data (Stage 1 training). (2) Hybrid training, which combines interpolated and non-interpolated data, and performs supervised and self-supervised training simultaneously. (3) Two-stage training, where GLP first reaches a local minimum loss using interpolated data and then undergoes domain-specific adjustments using SSL.


Regardless of the employed training approaches, the data were randomly divided into training and testing datasets in an 80:20 ratio. All training processes utilized the 5-fold cross-validation technique, and the reported results represent the mean value obtained from five repetitions of the training process \cite{CV_1995study}. Ablation studies were conducted to analyze the performance of different training processes combined with various interpolation methods. The outcome of the model was assessed using the R-squared ($R^{2}$) metric, which indicates the proportion of variance in the dependent variable that can be predicted by the independent variables in the model. $R^{2}$ values range from zero to one, with a value of one representing a perfect fit to the data and a value of zero indicating a poor fit. $R^{2} < 0$ suggests that the model performs worse than a horizontal mean line passing through the mean value of the data. Statistical significance to determine differences in model performance was assessed using an independent T-test, with \emph{p} $<$ 0.05 indicating statistical significance.

\subsection*{Downstream classifier design and validation}
The collection of the six laboratory values occurred at the time of performing PCI. Subsequently, $g$ (month-based) was computed based on the temporal disparity between the PCI and TVR dates. Patient information, including gender, age, and the six laboratory parameters, was collected and normalized following the aforementioned procedures. Due to the imbalanced distribution of patients between those with TVR occurrence (42) and those without (441), TVR-negative cases were randomly downsampled. As a result, only 84 patients entered the training process. Patient data are processed by the frozen GLPs, which are concatenated in a multimodal fashion (as shown in Figure~\ref{Training_flow}). 

Owing to the limited patient volume, non-neural network algorithms were chosen for training the downstream classifier. The included methods are Light Gradient Boosting Machine (LightGBM), Support Vector Machine (SVM), Logistic Regression (LR), and K-Nearest Neighbors (KNN). These approaches were selected for their diverse mechanisms. Cohen's Kappa was calculated to assess the agreements between the classifiers, and the mean value of Cohen's Kappa represented the overall agreements among the pairwise classifiers. A value of Kappa $\leq$ 0 indicates no agreement, while an increase in Kappa signifies an increase in agreement, with a value of 1 indicating perfect agreement.


Additionally, we compared the latent progress representations produced by GLP by extracting the outputs of $LIBC$ ($Progress_{emb}$) and $regressor$ ($Progress_{out}$), depicted in Figure~\ref{fig_model_design}. The transferred results based on the original data (normalized but not processed with GLP), $Progress_{emb}$, and $Progress_{out}$ were compared. The reported performance represents the mean value obtained after executing the training process five times. The downstream classifier is a binary classifier that distinguishes TVR occurrence (positive/negative). The evaluation metrics for TVR incidence include the area under the receiver operating characteristic curve (AUROC), accuracy, sensitivity, specificity, precision, and F1 score. To determine the statistical significance of the differences in evaluation metrics between the original data and the extracted representations, an independent T-test was also conducted. 

Furthermore, we plotted the distribution of the original data, $Progress_{out}$ at $g/2$, and $Progress_{out}$ at $g$ to visually depict the contribution of GLP throughout the process. Here, $g/2$ signifies that the iteration before reaching the intended event.

\subsection*{Patient recruitment and datasets}
Two datasets were obtained from two diverse medical institutes. The pretext dataset was obtained from the Chang Gung Research Database \cite{CGRD}, a multi-institutional electronic medical records database comprising original medical records of seven medical institutes in Taiwan. We included patients who were diagnosed with HTN prior to DM. Patients diagnosed with hypertension before the age of 40, those with any oncology visits, or individuals with observations spanning less than a year were excluded. The date of diagnosis was determined based on the International Classification of Diseases (ICD) encoding or the date of medication prescription. The ICD codes and medications used for the indications can be found in Appendix~\ref{definitions}. When a patient had been coded as having HTN or DM twice or more during a year, the onset date of the disease was defined as the first coded date. If the date of the first medication prescription preceded the ICD coded date, then the earlier date was designated as the onset date of the disease. The time interval between the HTN and DM diagnoses was set to be $\geq$ 3 months. 

We gathered data on patients between the onset of HTN and DM; that is, the patient was diagnosed with HTN, but yet to be determined as DM. Demographic information and laboratory data of the enrolled patients were collected, including age, sex, Chol/HDL-c, LDL-c, LDL-c/HDL-c, glucose AC, WBC, and UA. Erroneous values such as "NA" or "." were excluded. A total of 9,720 patients were included, and laboratory data were collected between January 2001 and December 2019.

The downstream dataset was acquired from the Taipei Veterans General Hospital, a tertiary hospital situated in northern Taiwan. We recruited 891 patients with noninvasive evidence of myocardial ischemia who underwent PCI between January 2005 and January 2022. Patients with ACS, acute decompensated congestive heart failure, acute or chronic infections, autoimmune diseases, malignancies with a prognosis of less than one year, unstable hemodynamic status, or those unable to receive dual antiplatelet therapy were excluded. Angiographically successful coronary intervention was defined as residual stenosis of less than 30\%, and coronary thrombolysis in myocardial infarction grade 3 flow was achieved at the conclusion of the procedure without any significant complications. All patients were followed up and to monitor the occurrence of TVR. 

Patients necessitating TVR were labeled as positive, and the corresponding dates were recorded. Patients without TVR until the end of January 2022 were classified as negative, with the date of January 31, 2022 recorded as the endpoint. Within this dataset, the available information encompassed age, sex, PCI date, TVR date, and the aforementioned six laboratory values (collected during the PCI procedure). The time intervals between the PCI and TVR dates were calculated. Patients with incomplete data were excluded from the analysis to avoid deviations into this study, as the imputation of missing values could potentially have that effect. Consequently, a total of 483 patients were included in the subsequent analysis, comprising 42 TVR and 441 non-TVR cases.

The pretext dataset are longitudinal observations that consists of multiple events, whereas the downstream dataset are episodic records that consists of one observation event for each patient. All patient data were de-identified prior to analysis. This study was approved by the Institutional Review Board of the Chang Gung Medical Foundation (No. 202000376B0) and Taipei Veterans General Hospital (No. 2019-12-012CC).

\section{Results}
\begin{figure}
\centerline{\includegraphics[width=\columnwidth]{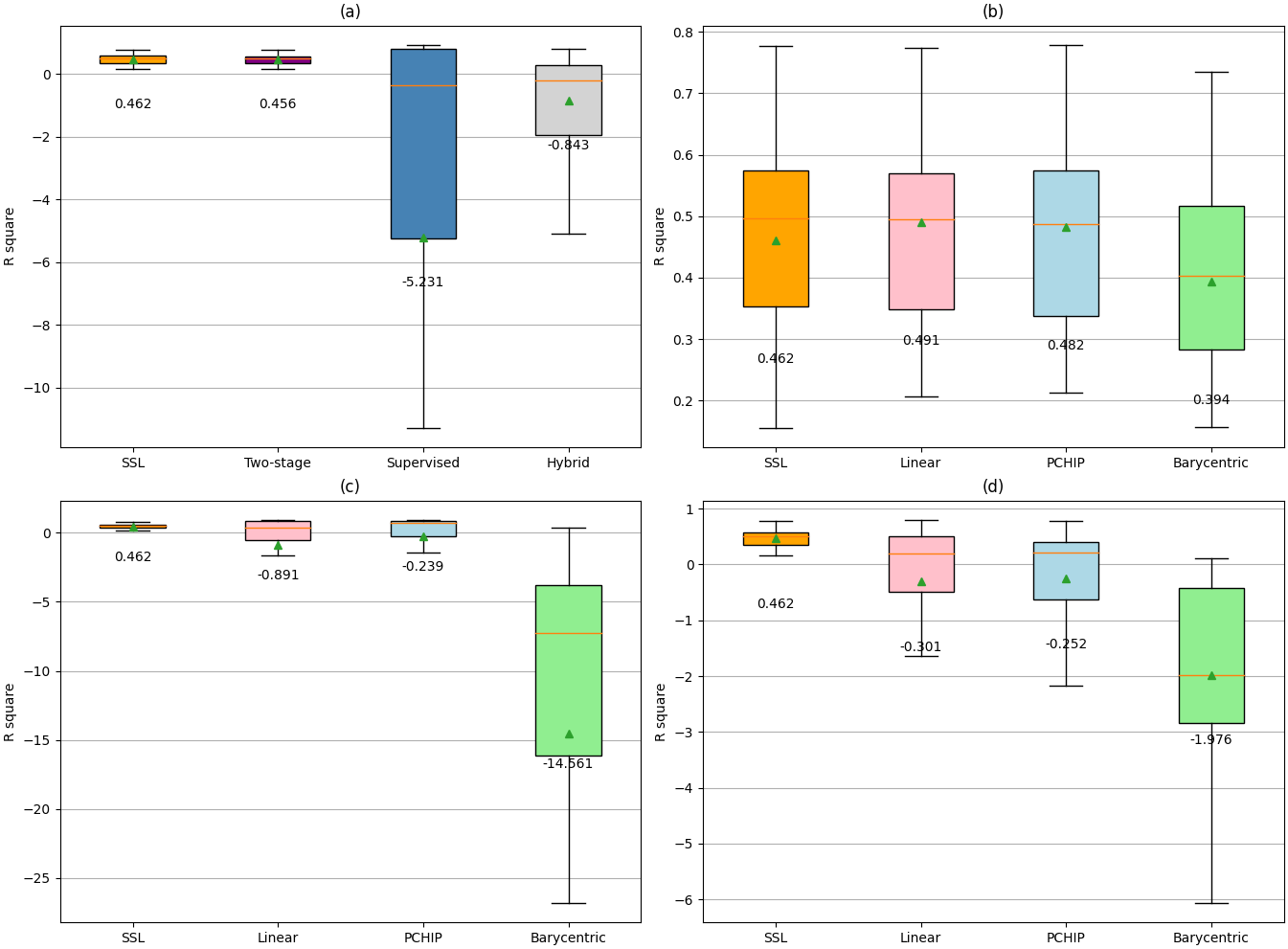}}
\caption{$R^{2}$ values for different training approaches. SSL (baseline) is compared with (a) averaged $R^{2}$ across different training approaches; (b) two-stage training employing different interpolation methods; (c) supervised training employing different interpolation methods; and (d) hybrid training employing different interpolation methods. The mean value is represented by the green triangle and its corresponding numerical figure is provided in the text below. The presented outcomes are an average of five repetitions of training and prediction, determined by aggregating the $R^{2}$ outcomes across $certain$ values ranging from 0 to 5. SSL: self-supervised learning.}
\label{fig_result}
\end{figure}

Table~\ref{table_demographic} illustrates the demographic information of the patients enlisted from the two datasets.  It is noteworthy that patients in the pretext dataset (HTN to DM patients) exhibit a relatively younger age compared to those in the downstream dataset (PCI patients). It is observed that TVR typically occurs within a period of 2.32 $\pm$ 2.64 years. Fig.~\ref{fig_result} depicted the overall performance of GLP, assessed by averaging the $R^{2}$ results across $certain$ ranging from 0 to 5. Fig.~\ref{fig_result}(a) provides an overview of the general disparities among different training approaches, incorporating all interpolation methods. It reveals that SSL and two-stage training exhibit similar performances, both achieving a mean $R^{2}$ value of 0.46 when rounded to the second decimal place. On the other hand, supervised training (Stage 1 training) and hybrid training achieved lower means with larger variations. 

Fig.~\ref{fig_result}(b) to Fig.~\ref{fig_result}(d) delves into the exploration of two-stage training, supervised training, hybrid training, respectively. All figures inspect into the effect of different interpolation methods. Observing from the figures, the two-stage training approach (Fig.~\ref{fig_result}(b)) is the only approach that surpasses SSL (baseline). Using linear ($R^{2}$ = 0.49, \emph{p} = 0.508) and PCHIP  ($R^{2}$ = 0.48, \emph{p} = 0.603) both yield better results than SSL, although the differences did not reach statistical significance. However, when compared to barycentric interpolation, both linear (\emph{p} = 0.031) and PCHIP (\emph{p} = 0.046) show significantly higher performance. Other training approaches, including supervised training depicted in Fig.~\ref{fig_result}(c) and hybrid training in Fig.~\ref{fig_result}(d), demonstrate weaker performance compared to the horizontal mean line ($R^{2} <$ 0), with larger variations.

\begin{table}
\caption{Demographic information of the recruit patients}
\label{table_demographic}
\setlength{\tabcolsep}{3pt}
\begin{tabular}{p{60pt}p{90pt}p{30pt}p{50pt}}
\toprule
\multicolumn{2}{c}{Items} & \multicolumn{2}{c}{Values}\\
\midrule
\multirow{6}{\linewidth}{HTN to DM patient \break(Pretext dataset)} 
& \multirow{2}{*}{Gender (n, \%)}& Male &5032	(0.518) \\
                       && Female & 4688	(0.482)\\
          & HTN onset age (mean, SD) & 56.965 & 9.653 \\
          & DM onset age (mean, SD)  & 61.884 & 9.878  \\
          & Duration between HTN and DM (year, mean, SD) & 4.992 & 3.945 \\
\midrule
\multirow{13}{\linewidth}{PCI patients \break (Downstream dataset)} 
& \multirow{2}{*}{Gender (n, \%)}& Male &402	(0.832) \\
                              && Female & 81 (0.168)\\
            & PCI treatement age \break (mean, SD) & 66.828 & 12.471 \\
& \multirow{2}{*}{TVR occurrence (n, \%)}& Yes &42 (0.087)\\
                              && No & 441 (0.913)\\
            & Duration between PCI and TVR occurrence \break (year, mean, SD) & 8.277 & 4.088 \\
            & Occurrence duration \break (year, mean, SD) & 2.319 & 2.640 \\
            & Non-occurred duration \break (year, mean, SD) & 8.844 & 3.734 \\
\bottomrule
\multicolumn{4}{p{251pt}}{n: number of samples; SD: Standard deviation; HTN: hypertension; DM: diabete mellitus; PCI: percutaneous coronary intervention; TVR: target vessel revascularization. Duration between HTN and DM signifies the cumulative observed timeframe for each individual. Occurrence duration delineates the temporal interval between PCI and TVR, and non-occurred duration characterizes the observed time span for patients who did not experienced TVR.} \\
\end{tabular}
\end{table}

\begin{table}
\centering
    \caption{$R^{2}$ of GLP based on optimized certainty mask configurations}
    \label{best_certainty}
    \begin{tabularx}{\columnwidth}{p{50pt}p{60pt}p{30pt}p{35pt}p{30pt}}
    \toprule
    & $certain$ & SSL & Two-stage & \emph{p-value} \\
    \midrule
    \multicolumn{5}{p{240pt}}{Linear}\\
    \midrule
    Chol/HDL-c  & 5 (2.4 months) & 0.509 & \B 0.602 &\\
    LDL-c       & 3 (4 months) & 0.373 & \B 0.376 &\\
    LDL-c/HDL-c & 3 (4 months) & 0.515 & \B 0.522 &\\
    Glucose AC  & 4 (3 months) & 0.648 & \B 0.666 &\\
    WBC         & 4 (3 months) & \B 0.778 & 0.774 &\\
    UA          & 4 (3 months) & 0.353 & \B 0.454 &\\
   \midrule
    Avg.        & 3.8 (3.1 months) & 0.529 & \B 0.566 & 0.773\\
    \bottomrule
    \multicolumn{5}{p{240pt}}{PCHIP}\\
    \midrule
    Chol/HDL-c  & 5 (2.4 months) & 0.392 & \B 0.586 &\\
    LDL-c       & 3 (4 months) & \B 0.367 & 0.341 &\\
    LDL-c/HDL-c & 2 (6 months) & 0.505 & \B 0.521 &\\
    Glucose AC  & 4 (3 months) & \B 0.644 & 0.602 &\\
    WBC         & 5 (2.4 months) & 0.751 & \B 0.779 &\\
    UA          & 5 (2.4 months) & 0.316 & \B 0.373 &\\
    \midrule
    Avg.      & 4 (3 months) & 0.496 & \B 0.534 & 0.702\\
    \bottomrule
    \multicolumn{5}{p{240pt}}{Chol/HDL-c: ratio of total cholesterol and high-density lipoprotein cholesterol; LDL-c: low-density lipoprotein cholesterol; LDL-c/HDL-c: ratio of low-density and high-density lipoprotein cholesterol; Glucose AC: fasting blood sugar; WBC: white blood cells; UA: Uric Acid. The differitation did not reach the level of statistical significance (\emph{p} $>$ 0.05). The month indicated in the brackets signifies the necessary follow-up visits for patients based on the requisite count of actual observations within a 12-month period.}\\
    \end{tabularx}
\end{table}

Table~\ref{best_certainty} provides a summary of the optimal $certain$ settings for each GLP, which were found to be inconsistent across different parameters. It is noteworthy that, in most parameters, the two-stage training process consistently outperformed SSL, although the degree of improvement remained marginal (\emph{p} $>$ 0.05). While specifying the $certain$ value resulted in enhancing GLP performance (evidenced by an increase from 0.49 to 0.57 for linear interpolation), the correlation between $R^{2}$ and $certain$ was determined to be weak based on Pearson correlation analysis (linear: -0.002, PCHIP: 0.026, and barycentric: -0.040). This result indicated that $R^{2}$ and $certain$ are not linear correlated. The detailed performances of $certain$ for each analysis can be found in Appendix~\ref{detail_r2}. Based the best-performing linear interpolation approach, we trained the GLP pretrain models by optimizing $certain$ value for each parameters individually.

\begin{table}
\centering
    \caption{Downstream task predictions}
    \label{non-nn}
    \begin{tabularx}{\columnwidth}{p{20pt}p{27pt}p{27pt}p{27pt}p{27pt}p{27pt}p{27pt}}
    \toprule
     & AUROC & Accuracy & Sensitivity & Specificity & Precision & F1\\
    \midrule
    &\multicolumn{6}{p{200pt}}{Original data (Mean Cohen's Kappa = 0.366)}\\
    \midrule
    LGBM & \B 0.602 & 0.671 & \B 0.717 & 0.632 & 0.587 & 0.645 \\
    SVM & 0.313 & \B 0.718 & 0.603 & 0.836 & \B 0.804 & \B 0.667 \\
    LR & 0.543 & 0.494 & 0.673 & 0.398 & 0.425 & 0.500 \\  
    KNN & \B 0.602 & 0.635 & 0.343 & \B 0.939 & 0.800 & 0.472 \\
    \midrule
    Avg. & 0.515 & 0.630 & 0.584 &	0.701 & 0.654 &	0.571 \\
    \midrule
    &\multicolumn{6}{p{200pt}}{$Progress_{emb}$ (Mean Cohen's Kappa = 0.307)}\\
    \midrule
    LGBM & \B 0.706 & \B 0.671 & \B 0.526 & \B 0.925 & \B 0.933 & \B 0.656 \\
    SVM & 0.601 & 0.435 & 0.450 & 0.511 & 0.231 & 0.290 \\
    LR & 0.536 & 0.494 & 0.362 & 0.657 & 0.510 & 0.357 \\  
    KNN & 0.394 & 0.518 & 0.142 & 0.830 & 0.267 & 0.183\\
    \midrule
    Avg. & 0.559 & 0.530* & 0.370* & 0.731 & 0.485 & 0.372* \\
    \midrule 
    &\multicolumn{6}{p{200pt}}{$Progress_{out}$ (Mean Cohen's Kappa = 0.941)} \\
    \midrule
    LGBM & \B 0.973 & \B 0.965 & \B 0.933 & 0.978 & 0.978 & \B 0.952 \\
    SVM & 0.907 & 0.882 & 0.767 & 0.969 & 0.933 & 0.822 \\
    LR & 0.863 & 0.871 & 0.708 & \B 1.000 & \B 1.000 & 0.824 \\
    KNN & 0.898 & 0.882 & 0.787 & 0.956 & 0.956 & 0.848 \\
    \midrule
    Avg. & \B 0.910** & \B 0.900** & \B 0.799** & \B 0.976** & \B 0.967** & \B 0.862** \\
    \bottomrule
    \multicolumn{7}{p{240pt}}{LGBM: Light Gradient Boosting Machine; SVM: Support Vector Machine; LR: Logistic Regression; KNN: K-Nearest Neighbors (KNN); AUROC: area under the receiver operating characteristics curve; Avg.: average; *: Statistical significant differences between $Progress_{emb}$ and original data; **: Statistical significant differences between $Progress_{out}$ and original data, as well as between $Progress_{out}$ and $Progress_{emb}$.}\\
    \end{tabularx}
\end{table}

\begin{figure}
\centerline{\includegraphics[width=\columnwidth]{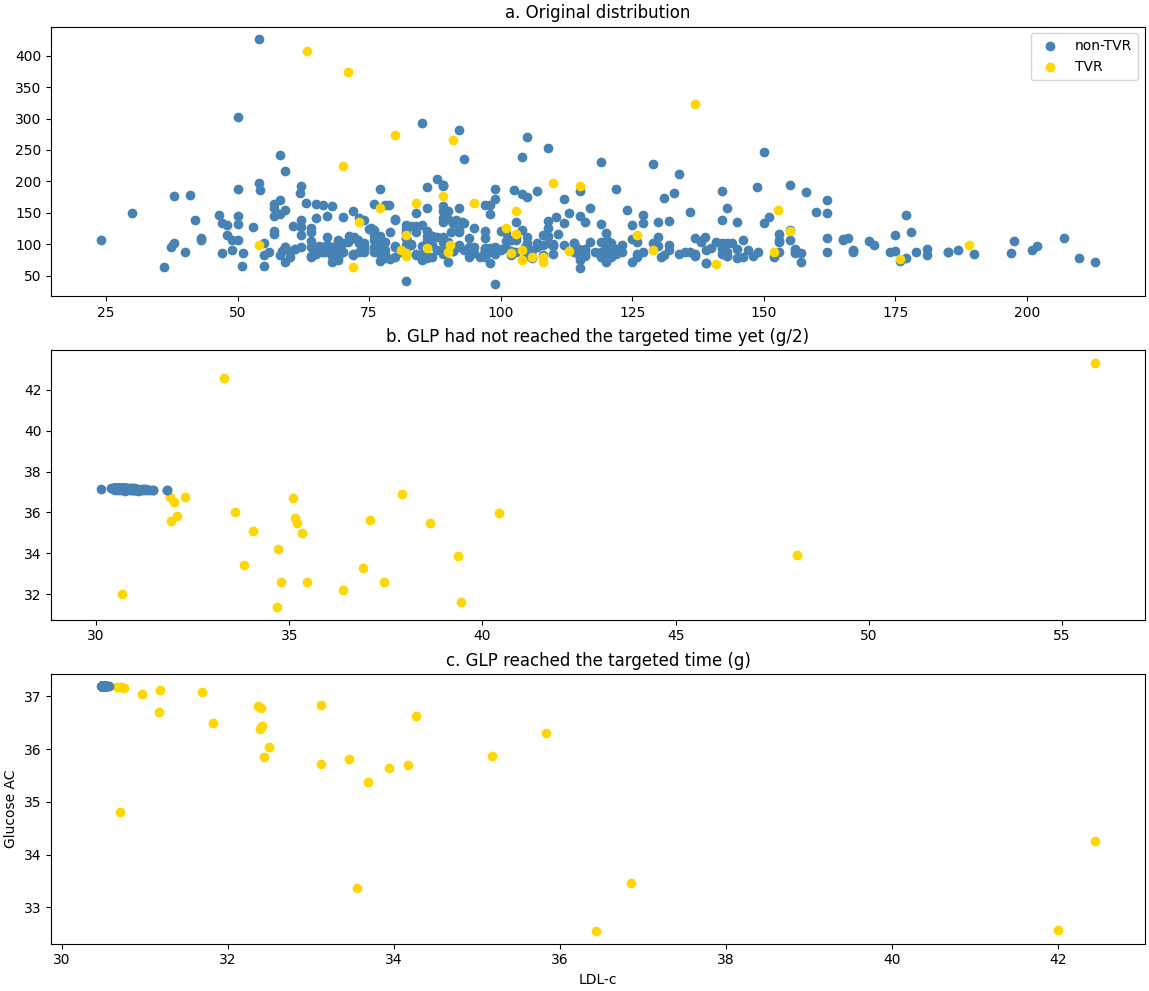}}
\caption{Alterations in distribution prior to and subsequent to GLP processing. The values were inverse from normalization. $g$: time gap between the input frame and prediction target.}
\label{GLP_output}
\end{figure}

Table~\ref{non-nn} presents the result of the downstream task. After the processing by GLP, $Progress_{out}$ exhibited a significant enhancement in classification performance. On average, the measures attained an AUROC of 0.91, accuracy of 0.90, sensitivity of 0.80, specificity of 0.98, and F1 score of 0.86. $Progress_{out}$ displayed a substantial superiority over the performance of the original data (\emph{p} $<$ 0.01) and $Progress_{emb}$ (\emph{p} $<$ 0.01). The key distinction between $Progress_{emb}$ and $Progress_{out}$ lies in the fact that $Progress_{out}$ represents a condensed version that solely indicates the laboratory values for the subsequent month, whereas $Progress_{emb}$ still retains information on age, gender, $certain$, discrete value range encoding, and laboratory values. This condensed version has a distillation effect, where the $regressor$ was trained to provide a more valuable indication of future trends based on these information.

Notably, LGBM emerged as the best-performing algorithm for both $Progress_{emb}$ and $Progress_{out}$, while SVM demonstrated the highest performance when utilizing the original data. The Kappa score demonstrates that the agreement among classifiers increased from 0.37 to 0.94, indicating that GLP alters the distribution of the original data and simplifies the classification task, irrespective of the algorithm mechanism employed.

Fig.\ref{GLP_output} visually illustrates the changes in data distribution, using glucose AC and LDL-c as exemplars. The figure depicts that prior to GLP processing (Fig.~\ref{GLP_output}a), distinguishing between TVR and non-TVR cases was challenging. However, after GLP processing, the non-TVR cases gradually converged towards a singular point, while the TVR cases displayed a more scattered distribution (as depicted in Fig.~\ref{GLP_output}b to c).

\section{Discussion}

We have successfully engage in translational engineering by transferring the progression of cardiovascular laboratory parameters from one patient group to another, without confining to episodic observations. To the best of our knowledge, this is the first study to apply the transfer of laboratory progression between patient groups, demonstrating that disease progression can be effectively transferred through deep neural network processing. The generality of our findings is assured. The datasets were sourced from two distinct medical institutions, signifying divergent patient populations, personnel, operational protocols, and measurement equipment. This finding opens up opportunities to leverage the trends observed in general cases for developing data-driven applications targeting patient groups with limited data availability.



Pretrained models capture the temporal dynamics and generate latent representations that enhance predictive capabilities for other tasks. With the widespread adoption of EHRs in modern hospitals, a large amount of data has been collected from prevalent cases, such as patients with HTN and DM. Conversely, specific cases, such as TVR occurance after PCI, remain limited in number. Given that HTN and DM are known risk factors for TVR \cite{braunwald1997shattuck}, we successfully transferred the trend of laboratory progress observed in HTN patients (who had yet to be diagnosed as diabetic) to predict the progression of PCI patients.

Prior investigations \cite{vincent2010stacked} have highlighted that training deep neural network models directly through gradient descent can yield randomly initialized models that are less optimized. In contrast, commencing the training process with a pretrained model enables the preservation and utilization of previously acquired knowledge, thereby transforming our randomly initialized models into exceptional pretrained feature extractors \cite{SSL_review}. The practice of acquiring knowledge from a more generalized domain and subsequently fine-tuning it to a specific domain during pretraining has previously exhibited efficacy \cite{lee2020biobert}. It has been observed that representations acquired from supervised objectives tend to be more domain-specific and possess limited transferability to out-of-distribution domains \cite{self_supervised_time_serious_data, SSL_review, SSL_ECG_2023}. Thus, modifying the model based on SSL confers enhanced flexibility and facilitates the extrapolation of knowledge across domains. This transformation enhanced our ability to infer patient progress. Despite the heterogeneous characteristics exhibited by the two patient cohorts, there existed intercorrelations among cardiometabolic diseases. 

SSL proves viable in bridging the gap caused by irregularity, absenteeism, and sparsity in prolonged monitoring, enabling adequate prediction. Learning from interpolated data in advance exhibits the capacity to enhance prediction performance, however, the effects vary across interpolation methods. Linear and PCHIP methodologies are designed to acquire a continuous function, while barycentric interpolation is tailored to leverage the center of mass for interpolation. The distribution pattern of laboratory progress resembles more closely a continuous curve that extends over time, rather than being clustered in discrete bundles. Our findings reveal that linear and PCHIP interpolation yield more informative estimations, enhancing the performance of SSL. In the case of LDL-c and LDL-c/HDL-c examinations, interpolation expands the tolerance for periods of absence. Thus, our work implies that with sustainable estimation, patient risks can be monitored with less frequent returns and examinations (beyond the conventional 3-month interval), alleviating the burdens of travel and medical expenses for patients.


The results indicate that without the support of GLP, the original data lacked sufficient distinctiveness to enable substantial TVR predictions. Upon processing the data with GLP, the classifier successfully achieved a satisfactory separation. Non-TVR cases gradually converged towards a single point, while TVR cases exhibited a more scattered pattern. These findings align with clinical observations, which suggest that stable patients exhibit less variation, whereas those with scattered observations face increased risks. The distinctiveness is not constrained by algorithmic mechanisms. However, it can be observed that the data were transformed from a hyperplane cluster distribution (where SVM demonstrated highest performance in the original distribution) to a more variable-specific representation (where LGBM outperformed other methods) and facilitating the creation of tree-based rules. Due to the limited size of the dataset, non-neural network algorithms were employed for TVR prediction, as deep neural networks did not perform well in this scenario.

\begin{figure}
\centerline{\includegraphics[width=\columnwidth]{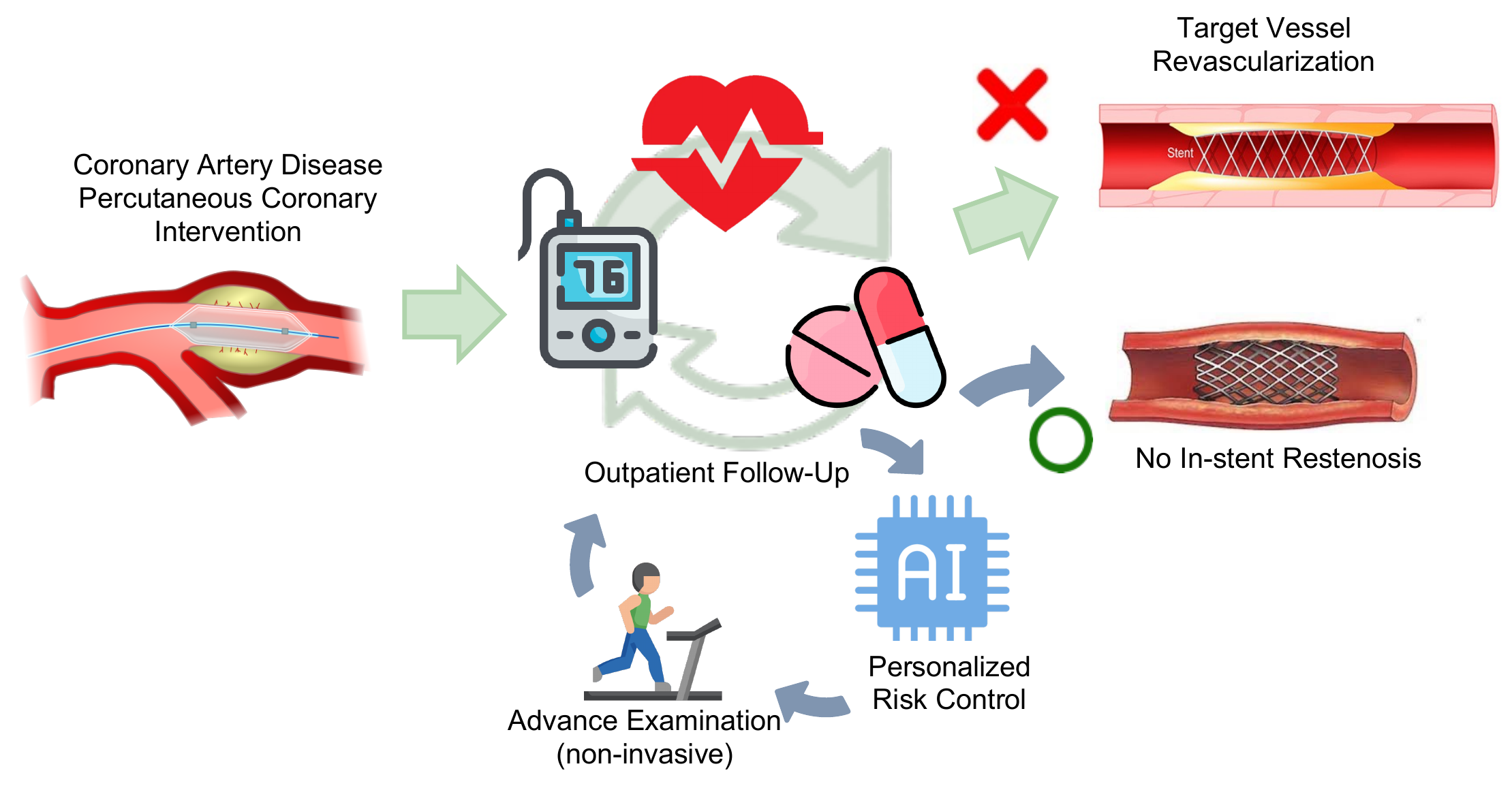}}
\caption{The user scenario of GLP adoption. Advanced examinations indicates treadmill test, thallium scan, or coronary computerized tomography.}
\label{AI_aided_workflow}
\end{figure}

The user scenario of GLP is depicted in Fig.~\ref{AI_aided_workflow}. Following PCI treatment, patients are required to undergo follow-up visits at the outpatient setting on a 3-month basis. Clinicians assess patient risk to prevent TVR occurrence. GLP identify patients at a higher risk, prompting the initiation of more advanced examinations, such as treadmill tests, thallium scans, or coronary computerized tomography (CT). Conversely, patients with a relatively lower restenosis risk can also be separated, serving as a screening tool for more precise event detection and examination resource allocation. In general, GLP optimizes treatment strategies and improves patient prognosis while maintaining simplicity and user-friendliness without relying on accumulating comprehensive TVR predictors.

To broader the applications to other disease, an effective approach for identifying the appropriate pretext and downstream tasks becomes indispensable. Our findings suggested that through leveraging the intercorrelations among diseases, such as comorbidity or shared risk factors, represents a more proficient approach for selecting pretext and downstream tasks. The intercorrelations imply underlying similarities and suggest the possibility of translating disease progression between different patient groups. This approach offers a more precise alternative to empirical trial-and-error methods.

This study is subjected to certain limitations: While other approaches, such as GAN, have demonstrated effectiveness in recovering the distribution of absent observations, we opted for interpolation as it is considered a simpler and less computationally expensive method that still achieves excellent transfer performances. However, further exploration is required to determine the necessity or indispensable advantages of integrating a more advanced distribution generator; Our work is constrained by the unavailability of genuine patient observations at the target event, which prevented us from precisely reversing the exact output. Additionally, it is important to acknowledge that, at present, the output of GLP corresponds to a representation that cannot be straightforwardly reconstructed as real-world values through simple inverse-normalization. To achieve accurate inverse mapping, training an additional network or a decoder might be necessary to effectively map the latent output to real-world values; Furthermore, it is worth noting that this study exclusively focused on the analysis of numeric laboratory results and its applicability is limited to other numeric examinations. To make GLP a more comprehensive laboratory information extractor, it is necessary to incorporate different types of laboratory analyses, including categorical variables.

SSL also inherits certain limitations, such as the need for a larger amount of training data, more training iterations, and computational expenses compared to supervised training \cite{breakingNewsSSL2022}. Additionally, deep neural networks employ a substantial number of trainable parameters and layers, making the model challenging to interpret \cite{XAI_survey, SSL_ECG_2023}. Thereby, SSL sacrificed interpretability for accuracy. This consideration should be taken seriously when utilizing the technology, as previous research has indicated that the level of trust in AI systems impacts the outcome of system utilization \cite{UI_design, adoption_factors}. Apart from interpretability, adopting AI system in to clinical workflow involved in multidisciplinary integration, encompassing areas such as the user interface design (research of Human-Computer Interactions) \cite{input_UI, UI_design}, and factors that influence the acceptance of technology \cite{adoption_factors}. Ensuring system usability and reliability while resolving the disparities between proof-of-concepts and real-life environments is crucial for the successful adoption of AI systems into daily practice. This endeavor necessitates further collaboration to facilitate the reform of healthcare.





\section{Conclusion}

Our research successfully translate the progression trends of cardiovascular laboratory parameters between patient groups by capitalizing on the advantages of SSL and pretrained models. This discovery paves the road for wider data-driven applications in healthcare, and also functions as a screening tool for more precise event detection and judicious allocation of examination resources. Additionally, our study suggests that patients can benefits from diminished frequency of visits and onerous examinations by implementing sustainable estimation. To accomplish healthcare reform through the utilization of AI systems, the key to success and optimal system utilization lies in the multifaceted aspects involved multidisciplinary collaborations. 


\vspace*{0.5 cm}

\section*{Acknowledgment}
This work was supported by the National Science and Technology Council, Taiwan, NSTC 111-2628-E-A49-026-MY3, NSTC 112-2321-B-075-002- and Taipei Veterans General Hospital under Grant V1118-031.

\pagebreak
\appendices 

\begin{appendices}
\section{Onset definition for patients with hypertension and diabetes mellitus.}
\begin{table}[!h]
\centering
    \label{definitions}
    \begin{tabularx}{\columnwidth}{p{3pt}p{40pt}p{180pt}}
    \toprule
    \multicolumn{2}{p{50pt}}{HTN ICD code}\\
    \midrule
    & ICD-9-CM & codes starting with 401, 402, 403, 404, 405\\
    & ICD-10   & codes starting with I10, I11, I12, I13, I15, I16\\
    \midrule
    \multicolumn{2}{p{50pt}}{HTN Drug}\\
    \midrule
    & \multicolumn{2}{p{230pt}}{Candesartan, Bisoprolol, Captopril, Enalapril, Amlodipine, Enalapril, Losartan Potassium, Carvedilol, Diltiazem HCL, Nifedipine, Felodipine, Lercanidipine, Nicardipine HCL, Nimodipine, Sotalol HCL, Verapamil HCL, Labetalol HCL, Fosinopril Sodium, Atenolol, Valsartan, Irbesartan, Metoprolol, Clonidine, Hydralazine HCL, Minoxidil, Benzyl Hydrochlorothiazide, Indapamide, Spironolactone, Pindolol, Nebivolol, Doxazosin SR, Eplerenone, Nicardipine, Ramipril, Imidapril HCL, Telmisartan, Olmesartan, Azilsartan Medoxomil.} \\
    \bottomrule
    \multicolumn{2}{p{50pt}}{DM ICD code}\\
    \midrule
    & ICD-9-CM & codes starting with 250\\
    & ICD-10   & codes starting with E08, E09, E10, E11, E13\\
    \midrule
    \multicolumn{2}{p{50pt}}{DM Drug}\\
    \midrule
    & \multicolumn{2}{p{230pt}}{Metformin} \\
    \bottomrule
    \multicolumn{3}{p{240pt}}{HTN: hypertension; DM: diabetes mellitus; ICD: International Classification of Diseases (ICD); CM: clinical modification.}\\
    \end{tabularx}
\end{table}


\section{Discrete value thresholds of laboratory parameters}
\begin{table}[!h]
\centering
    \label{lab_descret}
    \begin{tabularx}{\columnwidth}{p{60pt}p{30pt}p{70pt}}
    \toprule
     Item & Encoding & Value range \\
    \midrule
    \multirow{2}{\linewidth}{Chol/HDL-c} & 0 & $\leq 5$    \\
                                       & 1 & $> 5$    \\
    \midrule
    \multirow{2}{\linewidth}{LDL-c} & 0 & $\leq 160$   \\
                                     & 1 & $> 160$ \\
    \midrule
    \multirow{2}{\linewidth}{LDL-c/HDL-c} & 0 & $\leq 3.5$  \\
                                      & 1 & $> 3.5$  \\
    \midrule
    \multirow{3}{\linewidth}{Glucose AC} & 0 & $\leq 100$   \\
                                         & 1 & $100 <$ value $\leq 125$      \\
                                         & 2 & $ < 125$ \\
    \midrule
    \multirow{3}{\linewidth}{WBC} & 0 & $ < 4$ \\
                                  & 1 &  $4 \leq$ value $< 9$ \\
                                  & 2 & $ \geq 9$  \\
    \midrule
    \multirow{3}{\linewidth}{UA} & 0 & $\leq 3.4$  \\
                                  & 1 & $3.4 <$ value $\leq 7 $    \\
                                  & 2 & $< 7$ \\
    \bottomrule
    \multicolumn{3}{p{240pt}}{Chol/HDL-c: ratio of total cholesterol and high-density lipoprotein cholesterol; LDL-c: low-density lipoprotein cholesterol; LDL-c/HDL-c: ratio of low-density and high-density lipoprotein cholesterol; Glucose AC: fasting blood sugar; WBC: white blood cells; UA: Uric Acid.}\\
    \end{tabularx}
\end{table}

\pagebreak

\section{$R^{2}$ values of each laboratory parameters based on linear and PCHIP interpolation in two-stage training}
\begin{table}[!h]
\centering
    \label{detail_r2}
    \begin{tabularx}{\columnwidth}{p{5pt}p{14pt}p{14pt}p{14pt}p{14pt}p{14pt}p{14pt}p{14pt}p{14pt}p{14pt}}
    \toprule
    \multirow{2}{\linewidth}{c} 
        & \multicolumn{3}{c}{Linear} 
        & \multicolumn{3}{c}{PCHIP} & \multicolumn{3}{c}{Barycentric} \\
    \cmidrule(lr){2-4} \cmidrule(lr){5-7} \cmidrule(lr){8-10}
    & mean & $v_{CI}$ & $v_{2stage}$ & mean & $v_{CI}$ & $v_{2stage}$ & mean & $v_{CI}$ & $v_{2stage}$\\
    \midrule
    &\multicolumn{9}{p{240pt}}{Chol/HDL-c}\\
    \midrule
    0 & 0.539 & 0.012 & 0.397  & 0.546 & 0.025 & 0.345 & 0.427  & 0.007 & \num{-0.238} \\
    1 & 0.565 & 0.024 & 0.368  & 0.552 & 0.058 & 0.343 & 0.486  & 0.018 & \num{-0.177} \\
    2 & 0.570 & 0.024 & 0.326  & 0.564 & 0.021 & 0.260 & 0.527  & 0.074 & \num{-0.686} \\
    3 & 0.586 & 0.043 & 0.348  & 0.575 & 0.054 & 0.309 & 0.490  & 0.025 & \num{-0.302} \\
    4 & 0.486 & 0.119 & 0.258  & 0.377 & 0.077 & 0.231 & 0.519  & 0.109 & \num{-7.851} \\
    \B 5 & \B 0.602 & \B 0.057 & \B 0.281  & \B 0.586 & \B 0.039 & \B 0.228 & 0.295  & 0.094 & \num{-5.464} \\
    \midrule
    &\multicolumn{9}{p{240pt}}{LDL-c}\\
    \midrule
    0 & 0.326 & 0.022 & 0.613  & 0.329 & 0.025 & 0.562 & 0.247  & 0.040 & \num{-0.130} \\
    1 & 0.348 & 0.034 & 0.598  & 0.338 & 0.018 & 0.568 & 0.257  & 0.040 & 0.182  \\
    2 & 0.326 & 0.221 & 0.618  & 0.344 & 0.019 & 0.472 & 0.299  & 0.018 & \num{-0.240} \\
    \B 3 & \B 0.376 & \B 0.011 & \B 0.568  & \B 0.341 & \B 0.033 & \B 0.573 & 0.284  & 0.051 & \num{-0.034} \\
    4 & 0.207 & 0.349 & 0.272  & 0.213 & 0.119 & 0.382 & \num{-0.475} & 0.251 & \num{-6.924} \\
    5 & 0.234 & 0.095 & 0.402  & 0.311 & 0.070 & 0.526 & 0.157  & 0.164 & \num{-4.875} \\
    \midrule
    &\multicolumn{9}{p{240pt}}{LDL-c/HDL-c}\\
    \midrule
    0 & 0.479 & 0.028 & 0.466  & 0.481 & 0.029 & 0.426 & 0.374  & 0.019 & \num{-0.070} \\
    1 & 0.505 & 0.074 & 0.446  & 0.503 & 0.033 & 0.407 & 0.417  & 0.014 & 0.050  \\
    \B 2 & 0.509 & 0.043 & 0.399  & \B 0.521 & \B 0.019 & \B 0.313 & 0.444  & 0.020 & \num{-0.271} \\
    \B 3 & \B 0.522 & \B 0.053 & \B 0.414  & 0.496 & 0.029 & 0.382 & 0.454  & 0.032 & \num{-0.026} \\
    4 & 0.488 & 0.413 & \num{-0.077} & 0.328 & 0.167 & 0.124 & 0.321  & 0.502 & \num{-8.539} \\
    5 & 0.297 & 0.122 & 0.433  & 0.513 & 0.022 & 0.272 & 0.290  & 0.106 & \num{-3.942} \\
    \midrule
    &\multicolumn{9}{p{240pt}}{Glucose AC}\\
    \midrule
    0 & 0.480 & 0.040 & 0.426  & 0.480 & 0.053 & 0.392 & 0.360  & 0.014 & \num{-1.987} \\
    1 & 0.495 & 0.034 & 0.373  & 0.483 & 0.056 & 0.330 & 0.387  & 0.041 & \num{-1.799} \\
    2 & 0.546 & 0.096 & 0.262  & 0.508 & 0.090 & 0.268 & 0.451  & 0.085 & \num{-2.099} \\
    3 & 0.500 & 0.031 & 0.361  & 0.487 & 0.083 & 0.306 & 0.403  & 0.032 & \num{-2.496} \\
    \B 4 & \B 0.666 & \B 0.319 & \B 0.075  & \B 0.602 & \B 0.343 & \B 0.184 & 0.516  & 0.121 & \num{-4.839} \\
    5 & 0.544 & 0.229 & 0.203  & 0.577 & 0.252 & 0.167 & 0.533  & 0.280 & \num{-3.504} \\
    \midrule
    &\multicolumn{9}{p{240pt}}{WBC}\\
    \midrule
    0 & 0.664 & 0.024 & 0.245  & 0.657 & 0.022 & 0.226 & 0.564  & 0.027 & \num{-8.366} \\
    1 & 0.712 & 0.037 & 0.170  & 0.708 & 0.011 & 0.135 & 0.617  & 0.018 & \num{-3.608} \\
    2 & 0.744 & 0.022 & 0.106  & 0.758 & 0.022 & 0.060 & 0.701  & 0.013 & \num{-2.309} \\
    3 & 0.743 & 0.018 & 0.117  & 0.721 & 0.041 & 0.097 & 0.670  & 0.019 & \num{-3.949} \\
    \B 4 & \B 0.774 & \B 0.019 & \B 0.058  & 0.761 & 0.012 & 0.051 & 0.716  & 0.026 & \num{-7.218} \\
    \B 5 & 0.763 & 0.020 & 0.085  & \B 0.779 & \B 0.016 & \B 0.010 & 0.734  & 0.036 & \num{-6.471} \\
    \midrule
    &\multicolumn{9}{p{240pt}}{UA}\\
    \midrule
    0 & 0.311 & 0.029 & 0.581  & 0.310 & 0.047 & 0.519 & 0.254  & 0.053 & \num{-1.894} \\
    1 & 0.333 & 0.038 & 0.539  & 0.325 & 0.047 & 0.472 & 0.256  & 0.079 & \num{-1.313} \\
    2 & 0.339 & 0.119 & 0.479  & 0.337 & 0.065 & 0.431 & 0.269  & 0.085 & \num{-1.872} \\
    3 & 0.355 & 0.040 & 0.521  & 0.328 & 0.033 & 0.415 & 0.257  & 0.058 & \num{-2.296} \\
    \B 4 & \B 0.454 & \B 0.179 & \B 0.290  & 0.254 & 0.303 & 0.170 & 0.415  & 0.260 & \num{-3.265} \\
    \B 5 & 0.272 & 0.231 & 0.235  & \B 0.373 & \B 0.089 & \B 0.433 & 0.283  & 0.149 & \num{-2.886} \\
    \midrule
    Avg. & \B0.491 & 0.091 & 0.340 & 0.482 & \B0.068 & \B0.316 & 0.394 & 0.083 & \num{-2.825}\\
    \bottomrule
    \multicolumn{10}{p{250pt}}{$v_{CI}$: boundary range of the 95\% confidence interval; $v_{2stage}$: prediction variation between stage 1 and 2; c: $certain$. The bold figures are the chosen best performed settings. }\\
    \end{tabularx}
\end{table}

\end{appendices}

%
%

\vspace*{ 1 cm}

\bibliographystyle{ieeetr}
\bibliography{references}
%

%
%
%
%
%
%
%
\end{document}